# Dynamic Hypergraph-Enhanced Prediction of Sequential Medical Visits

Wangying Yang*
University of Southern California
Los Angeles, USA

Zitao Zheng
Independent Researcher
New Jersey, USA

Zhizhong Wu
University of California, Berkeley
Berkeley, USA

Bo Zhang
Texas Tech University
Lubbock, USA

Yuanfang Yang
Southern Methodist University
Dallas, USA

*Abstract*—This study introduces a pioneering Dynamic Hypergraph Networks (DHCE) model designed to predict future medical diagnoses from electronic health records with enhanced accuracy. The DHCE model innovates by identifying and differentiating acute and chronic diseases within a patient's visit history, constructing dynamic hypergraphs that capture the complex, high-order interactions between diseases. It surpasses traditional recurrent neural networks and graph neural networks by effectively integrating clinical event data, reflected through medical language model-assisted encoding, into a robust patient representation. Through extensive experiments on two benchmark datasets, MIMIC-III and MIMIC-IV, the DHCE model exhibits superior performance, significantly outpacing established baseline models in the precision of sequential diagnosis prediction.

## I. INTRODUCTION

In the prediction of sequential medical visits, the main objective of this paper is to predict the diseases that will be diagnosed during the patient's next visit based on their electronic health records. The patient's visit records detail various visit information and are arranged in chronological order. Many research efforts have focused on the temporal information of patient visit records and have achieved excellent performance. Among them, methods based on recurrent neural networks and graph neural networks have achieved significant results in the field of sequential visit prediction.

For methods based on recurrent neural networks [1], the key to their advantage lies in effectively capturing the temporal dependencies in the patient's visit sequence. However, the diseases between adjacent visits may be independent. For example, the disease in the current visit may be related to the disease in a distant visit, rather than the disease in the most recent visit.

In addition, existing methods based on recurrent neural networks for sequential visit prediction usually aggregate the representation of all diseases diagnosed in a single visit as the representation of that visit, which cannot model the interactions between diseases diagnosed in a single visit. Therefore, models based on recurrent neural networks perform poorly in the task of sequential visit prediction.

Recently, graph Neural Networks (GNNs) [2] have shown excellent performance in various fields such computer vision for interpreting non-Euclidean data like medical imaging[3-6], natural language processing for tasks like semantic parsing[7-8], and fraud detection in financial networks[9-11]. Unlike methods based on recurrent neural networks, methods based on graph neural networks construct a pair-wise graph structure for diseases diagnosed in a single visit. In real medical scenarios, diseases diagnosed in a single visit may be long-standing or may occur suddenly.

For predicting the diseases that will be diagnosed in the patient's next visit, the impact of chronic and acute diseases is obviously different; therefore, the fine-grained distinction between chronic and acute diseases is of great significance for improving the effectiveness of disease prediction.

Moreover, a disease is often induced by a variety of historical diseases. For example, during a visit, a patient may be diagnosed with diabetes, hypertension, and cardiovascular diseases at the same time; diabetes can increase the risk of heart disease and hypertension, hypertension can lead to the deterioration of heart disease, and is also a common cause of diabetes complications. Obviously, there are complex many-to-many or many-to-one relationships between diseases; therefore, modeling the high-order relationships of diseases diagnosed in a single visit at a fine granularity also has an important impact on the effectiveness of disease prediction. Electronic medical records contain a large amount of clinical event information, such as medical history and examination results, which can effectively reflect the patient's health status. However, existing methods often ignore this clinical event information.

To address the above issues, this paper proposes a disease prediction method based on dynamic hypergraph [12] networks (DHCE). This method first classifies diseases into acute and chronic diseases, and constructs dynamic hypergraphs to capture high-order relationships.

## II. BACKGROUND

The research on disease prediction and the application of deep learning techniques in medical diagnosis have seen significant advancements. This section reviews key contributions in these areas, emphasizing methodologies that align closely with the development and application of our Dynamic Hypergraph Networks (DHCE) model.

Lin et al. [13] presented a comprehensive study on constructing disease prediction models using AI and deep learning. Their work laid the foundation for understanding how computer-based learning techniques can predict diseases by analyzing patterns in electronic health records (EHRs). This aligns with our approach of leveraging dynamic hypergraphs to enhance prediction accuracy by identifying complex disease interactions. Zhan et al. [14] advanced feature extraction and recognition in medical imaging systems through deep learning techniques. This work highlights the importance of accurate feature extraction, a principle that is also critical in our DHCE model, particularly when encoding clinical event data from EHRs. Wang et al. [15] explored multimodal deep learning architectures for image-text matching, which underscores the necessity of integrating diverse data types. Similarly, our model integrates clinical text data with visit records to form a comprehensive patient representation. Gao et al. [16] proposed an enhanced encoder-decoder network architecture to reduce information loss in image semantic segmentation. The principles of minimizing information loss and maintaining high fidelity in data representation are integral to the DHCE model's design, particularly in capturing the nuanced relationships between diseases. Liu et al. [17] and Yang et al. [18] both examined the application of multimodal fusion deep learning models in disease recognition and diagnosis. Their research supports the notion that combining multiple data sources can significantly improve model performance, a strategy we employ in our DHCE model to enhance the prediction of sequential medical visits.

Graph Neural Networks (GNNs) have shown promise in various fields including medical diagnosis. Unlike traditional neural networks, GNNs effectively handle non-Euclidean data structures, which is crucial in understanding the relationships between different medical conditions diagnosed in a single visit. Xu et al. [19] discussed the advancements in medical diagnostics facilitated by deep learning and data preprocessing. Their work underscores the importance of preprocessing and accurately modeling data, which are critical components of our approach in constructing dynamic hypergraphs for disease prediction. Research by Xiao et al. [20] and Hu et al. [21] on feature extraction and early warning models for specific diseases, such as brain diseases and cardiovascular conditions, demonstrate the practical applications of deep learning in improving diagnostic accuracy and early detection. These studies reinforce the potential of our DHCE model to be adapted for specific disease prediction tasks by accurately capturing high-order interactions.

Additionally, Yan et al. [22] explored the use of neural networks in survival prediction across diverse cancer types, emphasizing the flexibility and robustness of these models in managing various medical conditions. This adaptability is mirrored in our approach, where the DHCE model can adjust to different datasets and disease types, as evidenced by our experiments on the MIMIC-III and MIMIC-IV datasets. In medical settings, patients may be diagnosed with multiple diseases during a single visit, which are typically represented by specific medical codes, such as ICD-9-CM or ICD-10. This coding not only helps in standardizing diagnoses but also in facilitating the application of GNNs to predict and analyze complex medical data effectively.

Given an electronic medical record dataset $\{\gamma_u \mid u \in \mathbb{U}\}$, where U denotes the set of patients, and $\gamma_u = (V_1^u, V_2^u, \cdots, V_T^u)$ represents the historical visit sequence of patient u. Each visit $V_t^u = \{C_t^u, E_t^u\}$ is recorded in the visit record with a subset of medical codes $C_t^u \subset \mathbb{C}$ and a subset of clinical events $E_t^u \subset \mathbb{E}$. $\mathbb{C} = \{c_1, c_2, \cdots, c_{|\mathbb{C}|}\}$ refers to the set of diseases in the electronic health record dataset, where diseases are represented by medical codes. Additionally, $\mathbb{E} = \{E_1, E_2, \cdots, E_{|\mathbb{E}|}\}$ is the set of clinical events in the electronic health record dataset, where $|\mathbb{E}|$ indicates the number of clinical events. For patient $u$, the i-th clinical event of the t-th visit $E_{iu}$Eiu consists of an event type $q_i \in \mathbb{Q}$ (e.g., laboratory tests, treatment prescriptions, surgeries) and a series of clinical event features $\{A_i^1, \cdots, A_i^{|m_i|}\}$, where $|m_i|$ denotes the number of clinical features. Each clinical feature $A_i^k$ is a tuple composed of a feature name $n_i^k \in \mathbb{N}$ and a feature value $v_i^k \in \mathbb{X}$, denoted as $(n_i^k, v_i^k)$, where $\mathbb{N}$ and $\mathbb{X}$ represent the set of feature names (e.g., disease codes, drug codes) and the set of feature values, respectively.

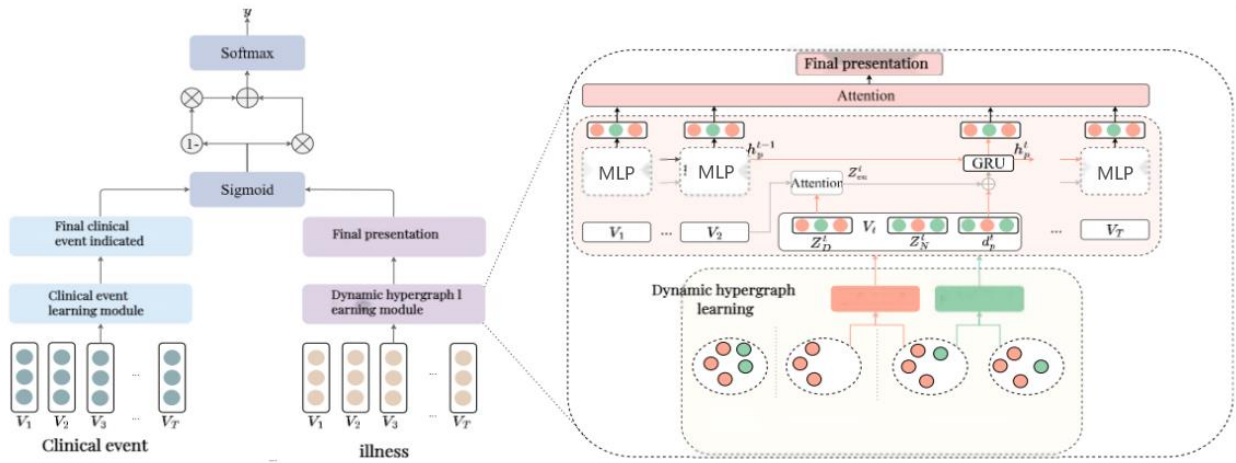

Figure 1. Model Structure

Problem Statement: Predict the diseases that will be diagnosed during the patient's $T+1$-th visit.

## III. METHOD

### A. Dynamic Hypergraph Learning Module

To capture the high-order interactions between diseases in a single visit record, this method introduces a hypergraph structure. Hypergraphs have applications in many fields, especially when dealing with complex network structures, as they can effectively describe high-order relationships between nodes, not just pairwise relationships.

In electronic medical record data, this section regards the diseases diagnosed in each visit by the patient as nodes, and the visit itself as a hyperedge, thereby constructing a hypergraph for a single visit. For example, as shown in the lower right part of Figure 1, the hypergraph of the t-th visit consists of 5 nodes and one hyperedge, where each node represents a disease diagnosed in the t-th visit, and the 5 nodes are connected by one hyperedge.

This method constructs a hypergraph for each visit. Since a patient has multiple visits, and the diseases diagnosed in each visit are not completely the same, a dynamic hypergraph for the patient's historical visits can be obtained. In this section, let $\mathcal{G} = \{\mathcal{G}^1, \mathcal{G}^2, \cdots, \mathcal{G}^r\}$ represent the dynamic hypergraph, and $\mathcal{G}^t = (\mathbb{P}^t, \mathbb{H}^c)$ represent the hypergraph of the patient's t-th visit. $\mathbb{P}' \subset \mathbb{C}$ represents the set of all nodes in the hypergraph $\mathcal{G}''$, that is, the set of all diseases diagnosed in the patient's t-th visit. Similarly, H' represents the edges of the hypergraph $\mathcal{G}'$. $\mathcal{G}'$ has an adjacency matrix O', the size of which is $|\mathbb{P}'|\times|\mathbb{H}'|$. When the patient u has disease $c_i$ in the t-th visit record, $\mathbf{O}_{c,u}^t$ is 1, otherwise it is 0.

Further divides these diseases into chronic and acute diseases:

Chronic diseases: $\mathbb{D}_p^t = \mathbb{D}^t \wedge \mathbb{D}^{t-1} \in \{0,1\}^{|\mathbb{C}|}$, that is, when a disease appears in the t-th visit record and also appears in the t-1 visit record.

Acute diseases: $\mathbb{D}_{em}' = \mathbb{D}^f \wedge \neg(\mathbb{D}^{-1}) \in \{0,1\}^{|\mathbb{C}|}$, that is, when a disease appears in the t-th visit record but does not appear in the t-1 visit record.

After dividing the diseases diagnosed in the t-th visit into two types, this section constructs two types of subgraphs based on the hypergraph $\mathcal{G}'$:

Local disease hypergraph $\mathcal{G}_D^t$: This is a hypergraph composed of chronic diseases diagnosed in the t-th visit, in which chronic diseases serve as nodes in the hypergraph and the t-th visit serves as the edge in the hypergraph.

Acute disease hypergraph $\mathcal{G}_{DN}^t$: This is a hypergraph that describes the relationship between each acute disease and chronic diseases. In this hypergraph, diseases serve as nodes in the hypergraph and the t-th visit is the edge in the hypergraph.

In real medical scenarios, diseases diagnosed in a single visit may have been long-standing or may have occurred suddenly. To capture these effects, in the dynamic hypergraph learning module, this section extracts local context and acute context.

Local context: For each diagnostic node (i.e., the node corresponding to the diagnosed disease in the hypergraph), this section aggregates the representations of diagnostic nodes connected to it in the hypergraph $\mathcal{G}_D'$, and aggregates the representations of acute disease nodes connected to it in the hypergraph $\mathcal{G}_{DN}^t$

Acute context: For each acute disease node, this section aggregates the representations of diagnostic nodes connected to it in the hypergraph $\mathcal{G}_{DN}'$, thereby obtaining the acute context.

The above operations capture the fine-grained high-order relationships of diseases in the same visit. Obviously, diseases in different visits also have certain relationships. To further capture the relationships between diseases in different visits, for the acute diseases diagnosed in a single visit $\mathbb{D}_{em}^t$, they are often induced by a variety of historical diseases. In this section, a click attention is designed as a transfer function to capture the acute transfer context, and the corresponding formula is as follows:

$$\mathbf{Z}_{en}^t = \text{Att}(\mathbf{Z}_N^{t-1}, \mathbf{Z}_N^{t-1}, \mathbf{Z}_D^t)$$

Using GRU to capture the interaction between the representation of chronic diseases $\mathbf{I}_p^t$ in the t-th visit and the acute transfer context $\mathbf{Z}_{cn}^t$ in the t-th visit, the transfer output of the t-th visit $\mathbf{Z}_p^t$ is obtained.

Afterward, this section uses a max-pooling operation on the transfer output to generate the visit representation $\mathbf{v}^t$ of the t-th visit.

Finally, this section uses an attention mechanism to calculate the influence of each historical visit on the prediction result, obtaining the final visit representation.

### B. Clinical Event Learning Module

Preprocessing: This section begins by converting clinical features involved in clinical events into textual descriptions. For a patient u's i-th clinical event in the t-th visit record, it is composed of the type of clinical event and the clinical features corresponding to $\{A_i^1, \cdots, A_i^{|m_i|}\}$. Each clinical feature $A_i^k$ consists of a feature name and its corresponding value, forming a tuple $(n_i^k, v_i^k)$. A clinical event in a single visit can be represented as:

$[\text{CLS}]q_i[\text{SEP}]n_i^1[\text{SEP}]v_i^1[\cdots[\text{SEP}]n_i^{|m_i|}[\text{SEP}]v_i^{|m_i|}$

, clinical event type, feature name, feature value, [SEP], where [CLS] and [SEP] are special tokens.

Clinical Event Representation: To capture information related to the patient's health status within clinical events, this section uses the Bio-Clinical BERT model to encode the clinical events, thereby gaining a better understanding of the patient's health condition. Specifically, the sequence of clinical events of type for patient u in the tth visit record can be formalized as:

$$\mathbf{m}_q = f\left(S(q_i), S(n_i^1), S(v_i^1), \ldots, S\left(n_i^{|m_i|}\right), S\left(v_i^{|m_i|}\right)\right)$$

Clinical Event Representation Aggregation: In each visit record, there will be a sequence of various types of clinical events. To obtain an overall representation of the clinical events, this section uses an attention mechanism to aggregate the representations of various types of clinical events, described as:

$$\mathbf{O}_e = \text{aggr}\left(\mathbf{m}_{q_1}, \mathbf{m}_{q_2}, \ldots, \mathbf{m}_{q_4}\right)$$

*C. Visit Prediction Learning Module*

To fully leverage the final visit representation and the final clinical event representation, this section utilizes a gating mechanism to fuse the two representations to generate a patient representation. A softmax function is used to predict the diseases that will appear in the patient's next visit. This section further optimizes the disease prediction model based on dynamic hypergraph networks proposed in this chapter using the cross-entropy loss function in formula.

$$F = sigmoid\left(\mathbf{W}_e \mathbf{O}_e + \mathbf{W}_v \mathbf{O}_v + \mathbf{b}_f\right)$$
$$\mathbf{u} = F \odot \mathbf{O}_v + (1 - F) \odot \mathbf{O}_e$$
$$\hat{y} = softmax\left(\mathbf{W}_y \mathbf{u} + \mathbf{b}_y\right)$$
$$\mathcal{L} = \frac{1}{T+1} \sum_{i=2}^{T} -(y_{\text{true}}^T \log \hat{y} + (1 - y_{\text{true}})^T \log(1 - \hat{y}))$$

In the formulas above, $\mathcal{L}$ represents the loss function; $\mathbf{O}_e$ represents the final clinical event representation; $\mathbf{O}_v$ represents the final visit representation; $\mathbf{W}_e$、$\mathbf{w}_r$、$\mathbf{W}_y$、$\mathbf{b}_f$, and $\mathbf{b}_y$ are learnable parameters; $\hat{y}$ is a multi-hot vector.

## IV. EXPERIMENT

*A. Experiment Settings*

*1) Dataset*

We utilized two benchmark datasets, namely MIMIC-III and MIMIC-IV [23], which are two publicly available electronic health record datasets created by the Laboratory of Computational Physiology at the Massachusetts Institute of Technology. MIMIC-III is the third version of the MIMIC dataset, containing medical information of approximately 40,000 patients from the period 2001 to 2012. MIMIC-IV is the latest version of the MIMIC dataset, containing medical information of about 80,000 patients from 2008 to 2019.

In the process of creating accessibility-linked data for analysis, we adopted the approach proposed by Li et al. [24]. Their method involves the use of publicly available accessibility datasets to enhance the interpretability and applicability of health data. By integrating these methodologies, we aim to enhance the robustness of our dataset, facilitating a deeper understanding of patient trajectories and outcomes.

*2) Baselines*

We compares the performance of the proposed DHCE model with the following state-of-the-art baseline models: Dipole [25], HiTANet [26], Deepr [27], GRAM [28], G-BERT [29].

*B. Experiment Results Analysis*

Table 1 Experiment Result Acc

| Algorithm | MIMIC-II | MIMIC-IV |
|---|---|---|
| Dipole | 19.35 | 24.98 |
| Timeline | 20.46 | 25.75 |
| HiTANet | 21.15 | 26.02 |
| GRAM | 21.52 | 26.51 |
| G-BERT | 19.88 | 25.86 |
| DHCE (our) | 24.24 | 29.53 |

Table 1 presents the experimental results conducted on two real datasets, including nine baseline models and the model proposed in this chapter. It is evident that on both datasets, DHCE achieved the best results on three metrics, thereby validating the effectiveness of the method proposed in this chapter. Based on the data in the table, the following conclusions are drawn:

Firstly, RNN-based methods (e.g., RETAIN, Dipole) outperform CNN-based methods (e.g., Deepr). CNN-based methods can only consider context information of a fixed size and cannot fully capture global, long-distance dependencies, while RNN-based methods fully leverage their advantage in modeling sequential data, being able to capture the temporal sequence dependencies between each visit in the medical records.

Secondly, RNN methods with attention mechanisms (e.g., Timeline, HiTANet) perform better than RNN methods without attention mechanisms. This is likely because the attention mechanism can effectively capture the importance of each visit and various clinical variables for the prediction outcome.

Thirdly, graph-based methods (e.g., GRAM, CGL, and Chet) outperform RNN-based methods in terms of performance. This may be because existing RNN-based sequential visit prediction models typically integrate the representations of diseases diagnosed in a single visit as the visit representation, and cannot effectively model the interaction relationships between diseases diagnosed in a single visit. This results in RNN-based models not achieving as good results as GNN-based models.

Fourth, the DHCE model proposed in this chapter significantly outperforms all baseline models in terms of performance. Although Chet can capture the clinical relationships between diseases diagnosed in a single visit, it ignores the fine-grained high-order information between diseases and does not fully utilize the clinical events in the electronic health records. Similarly, although RNN methods with attention mechanisms can select important clinical features in the medical records, they cannot effectively utilize the clinical events in the visit records, nor can they capture the fine-grained high-order interactive features between diagnosed diseases. Therefore, the method proposed in this chapter demonstrates better performance than these baseline models.

## V. CONCLUSION

This paper addresses limitations in sequential diagnostic prediction by introducing a dynamic hypergraph network-based method. It categorizes single-visit diagnosed diseases into chronic and acute types, creating hypergraphs to model their high-order interactions for visit representation. Clinical features from electronic health records are textually described and encoded via a medical language model to form clinical event representations. These are merged with visit data into a patient representation, enhancing the prediction of subsequent visit diagnoses. Experiments on two datasets reveal the model's marked superiority over baseline approaches.